\pdfoutput=1

\documentclass[11pt]{article}

\usepackage{ACL2023}

\usepackage{times}
\usepackage{latexsym}

\usepackage[T1]{fontenc}

\usepackage[utf8]{inputenc}

\usepackage{microtype}

\usepackage{inconsolata}

\usepackage{graphicx}
\usepackage{caption}
\usepackage{subcaption}
\usepackage{natbib}
\usepackage{mwe}
\usepackage{booktabs}
\usepackage{multirow}
\title{Towards Agile Text Classifiers for Everyone}

\author{Maximilian Mozes$^{1,2}$\thanks{\quad Equal contribution.}\,\,\thanks{\quad Work done during an internship at Google Research.}\quad Jessica Hoffmann$^{1\ast}$\quad Katrin Tomanek$^1$\quad Muhamed Kouate$^{1\dagger}$ \\\textbf{Nithum Thain$^1$\quad Ann Yuan$^1$\quad Tolga Bolukbasi$^1$\quad Lucas Dixon$^1$}\\
$^1$Google Research\\
$^2$University College London\\
\small{\{\texttt{jhoffmann,katrintomanek,kouate,nthain,annyuan,tolgab,ldixon\}@google.com}}\\
\small{\texttt{maximilian.mozes@ucl.ac.uk}}
}

\begin{document}

\newcommand{\tfivexxl}{T5 XXL}
\newcommand{\palm}{PaLM 62B}
\newcommand{\perspective}{PerspectiveAPI}
\newcommand{\parlai}{ParlAI}
\newcommand{\parlaistandard}{ParlAI Single Standard}
\newcommand{\parlaiadversarial}{ParlAI Single Adversarial}
\newcommand{\parlaimulti}{ParlAI Multi}
\newcommand{\badtwo}{BAD-2}
\newcommand{\badfour}{BAD-4}
\newcommand{\bftab}{\fontseries{b}\selectfont}

\maketitle

\begin{abstract}
Text-based safety classifiers are widely used for content moderation and increasingly to tune generative language model behavior---a topic of growing concern for the safety of digital assistants and chatbots. However, different policies require different classifiers, and safety policies themselves improve from iteration and adaptation. This paper introduces and evaluates methods for agile text classification, whereby classifiers are trained using small, targeted datasets that can be quickly developed for a particular policy. Experimenting with 7 datasets from three safety-related domains, comprising 15 annotation schemes, led to our key finding: prompt-tuning large language models, like \palm{}, with a labeled dataset of as few as 80 examples can achieve state-of-the-art performance. 
We argue that this enables a paradigm shift for text classification, especially for models supporting safer online discourse. Instead of collecting millions of examples to attempt to create universal safety classifiers over months or years, classifiers could be tuned using small datasets, created by individuals or small organizations, tailored for specific use cases, and iterated on and adapted in the time-span of a day.
\end{abstract}

\section{Introduction}

\begin{figure}[!ht]
     \centering
     \includegraphics[width=\columnwidth]{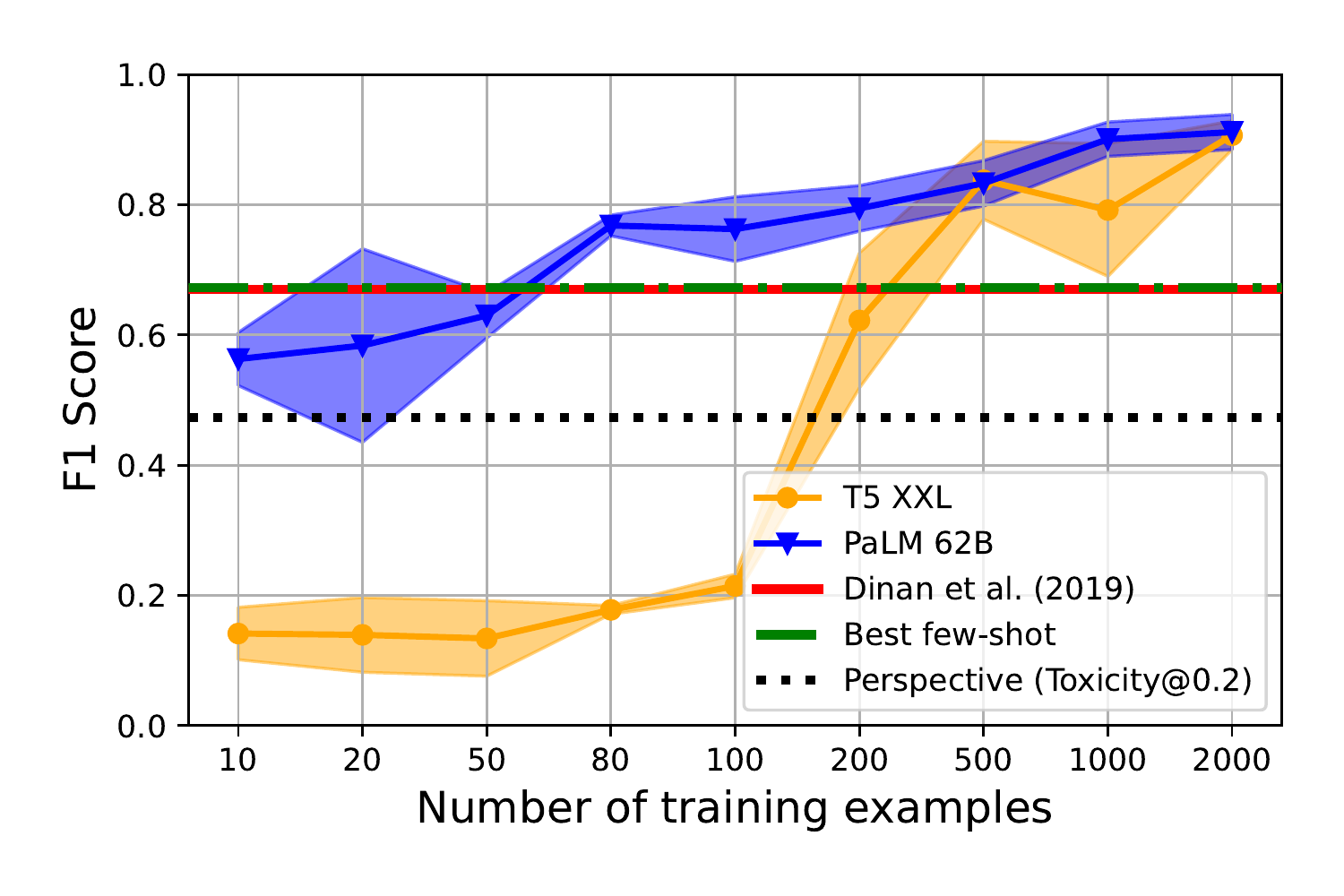}
    \caption{Prompt-tuning \palm{} and \tfivexxl{} with as few as 80 and 500 examples, respectively, outperforms both in-context learning (12-shot) on \palm{} and a BERT model fine-tuned on 24,000 training examples as reported in~\citet{dinan-etal-2019-build} for the \parlaiadversarial{} dataset.}
    \label{fig:prompt_tuning_intro_figure}
\end{figure}

Conversation moderation has changed rapidly over the past decade as platforms have evolved new tools. In the last few years, general purpose classifiers supporting online discourse, like Perspective API, have seen broad adoption; they are used to assist moderation~\cite{doi:10.1177/20539517211046181}, give feedback to authors~\cite{open-web2020}, and advance research in online safety. The Perspective API's most widely used model is for toxicity detection and was trained on hundreds of millions of annotations~\cite{10.1145/3308560.3317593}. This results in a useful model for its domain distribution. However, the way people use language is continuously changing, many forums have different policies, and policies themselves change frequently (e.g., to deal with new topics, such as COVID-19). In practice, one faces the challenge of either using the model with the quality degradation caused by the distribution and policy shift, or of training a custom model. Training a new high-quality neural text classification model (i.e., fine-tuning models such as BERT~\cite{devlin-etal-2019-bert}) typically requires collecting thousands or millions of textual annotations, a process which is both time-consuming and cost-intensive. 

In parallel to these developments, there has been rapid progress in chatbots, with ChatGPT representing a particular turning point in public awareness of the capabilities of large language models (LLMs).\footnote{\url{https://openai.com/blog/chatgpt/}} An important emerging strand of this research explores the role of human feedback and reinforcement learning in mitigating safety concerns with their outputs~\cite{AnthropicBai2022,DeepMindSparrowGlaese2022}. This field has not yet arrived at consensus on policies, nor developed high-quality large datasets for the proposed policies. Classifiers are already used to scale human feedback for tuning models, but being able to quickly iterate on high-quality text classifiers could play a particularly important role for the safety of modern chatbots.

This paper explores alternative approaches to text classification that leverage modern generative large language models like T5~\cite{raffel2020exploring} and GPT-3~\cite{brown2020language}. Not only can LLMs generate comments and conversations, they can also themselves act as safety classifiers, detecting which comments may need moderation \cite{chaudhary2021countering,rieder2021fabrics}. We explore prompting these models in two ways: using few-shot examples---often referred to as {\em in-context learning} (ICL), which includes task demonstrations in the input prompt and does not require model parameter updates ---and \textit{parameter-efficient tuning} (PET), a branch of transfer learning that adapts only a small number of parameters for new tasks and has been shown to obtain performance comparable to fine-tuning all parameters~\cite{li-liang-2021-prefix, lester-etal-2021-power, vu-etal-2022-spot}. Recent work suggests that PET may also enable more data-efficient training of models~\cite{liu2022few,agrawal2022qameleon}.

This paper's key contribution is to show the surprising effectiveness of PET for text classification on small datasets related to having safe and productive online discourse (Figure~\ref{fig:prompt_tuning_intro_figure}). A novelty of our study is the scale of data we experiment with: in addition to few-shot templates that are widely studied in the LLM literature, we explore datasets too large to fit into an LLM's context window for prompting, but too small for traditional fine-tuning.\footnote{We noticed that fine-tuning LLMs on small-scale datasets can lead to overfitting.}
Our experiments use two recent LLMs, \tfivexxl{}~\cite{raffel2020exploring} and \palm{}~\cite{chowdhery2022palm}, on three related domains. The first domain concerns offensiveness in dialogue, for which we study 5 datasets: three from \parlai{}~\cite{dinan-etal-2019-build} and two from Bot Adversarial Dialogue~\cite{xu2021bot}. The second domain concerns 7 different annotation schemes for attributes related to toxicity in online comments that were introduced in the Unhealthy Comment Corpus~\cite{price-etal-2020-six}. For the third domain, we evaluate three new annotation schemes inspired by Wikipedia's neutral point of view (NPOV) on a new dataset we created.\footnote{\url{https://en.wikipedia.org/wiki/Wikipedia:Neutral_point_of_view}} We focus specifically on such moderation tasks to demonstrate the effectiveness of PET in the context of small datasets, however the methods provided in this work are adaptable to any text classification task.

Our primary result is that prompt-tuning provides high performance ($>0.9$ ROC-AUC) for text classifiers across all domains we studied, matching or exceeding the quality of neural models on the same domain that were trained on up to 500x more annotated data, while also requiring only a fraction of the parameters (up to 1300x less). We also found our methods significantly surpass widely used classifiers trained on tens of millions of annotations on closely related domains. We argue this represents a new paradigm for training agile, domain-adapted safety classifiers using PET for LLMs.

\section{Related work}

\subsection{Safe online dialogue}

Over the past decade, a significant branch of research has been motivated by the goal of improving the safety of online conversations. 
A popular approach is developing classifiers to detect whether individual comments contain toxicity, personal attacks, hate speech, and other unhealthy attributes of online conversations \cite{wulczyn2017ex, davidson2017automated}. Classifiers for positive attributes such as constructiveness \cite{kolhatkar2020classifying} have also been explored.
The underlying models for detecting these attributes have also grown in sophistication, from linear models such as support vector machines~\cite{macavaney2019hate} to more modern approaches that rely on deep learning architectures like CNNs~\cite{gamback-sikdar-2017-using} and, more recently, transformers \cite{caselli2020hatebert, zhou2021challenges}.

With the rapid progress of chatbots, online chat support, and digital assistants, there has been a growing focus on multi-turn dialogue, with the goal of improving conversational agents and making them more robust to adversarial users \cite{dinan-etal-2019-build, xu2021bot}. Another recent area of research has focused on reducing the toxicity generated by large language models, although this risks diminished performance for the language of marginalized groups \cite{welbl2021challenges}.

Our contribution explores 9 ratings schemes across 7 datasets related to classifications of negative and positive characteristics of multi-turn dialogue and online commentary.

\subsection{Large language models}

Large language models, also known as foundation models, use the transformer architecture~\cite{vaswani2017attention}, typically have tens to hundreds of billions of parameters  and are pretrained on datasets consisting of hundreds of billions of tokens. These pretrained models, when fine-tuned, have demonstrated state-of-the-art performance on a multitude of tasks \cite{devlin-etal-2019-bert,chowdhery2022palm,tran2022plex}. 
One of the most important characteristics of LLMs is that they can be prompted with input texts to configure them to perform new tasks~\cite{radford2019language, brown2020language}. This ability to use text in the input to solve new tasks is called {\em in-context learning} (ICL). The prompting method that has the highest performance and which can directly use a small fragment of an existing task dataset is called {\em few-shot prompting}. This involves inserting a few examples of the task (typically between 2 and 20 examples) into the LLM prompt before the new example input. There is now a significant branch of research on improving ICL performance using, for example, self-consistency \cite{wang2022self} or chain-of-thought reasoning \cite{wei2022chain}. 

Our contribution focuses on evaluating the performance of few-shot prompts, leaving further comparisons to the quickly growing catalog of prompting tricks as further work.

\subsection{Parameter-efficient tuning}

Domain adaptation for pretrained language models has typically achieved state-of-the-art performance by fine-tuning all model parameters on relatively large datasets~\cite[tens of thousands to millions of annotations;][]{peters2018deep,devlin-etal-2019-bert}. However, a recent branch of research has found that one can keep the majority of the parameters in a large pretrained LLM fixed, instead updating only a small fraction of the parameters, with comparable downstream task performance~\cite{li2021prefix, lester-etal-2021-power}. This branch of research has largely focused on the trade-offs between how many parameters to tune, which parameters to tune (including the addition of new parameters), how to initialize them~\cite{vu2022spot}, and the resulting performance. However, a few new studies have indicated that these methods can also perform better than ICL when given the same tiny set of examples~\cite{liu2022few,agrawal2022qameleon}. 

We specifically focus on prompt-tuning, which prepends a set of learnable token embeddings (also called \textit{soft prompts}) to an LLM's input via concatenation. The soft prompt vector is optimized during training using the cross-entropy loss, while the LLM parameters remain fixed. At inference time, the trained token embeddings are then fed into the model along with the input prompt. Prompt-tuning is one of the most parameter-efficient approaches to PET; this allows the LLM to remain fixed and the soft prompt to be provided at query time with the input. In particular, it avoids having to manage multiple copies of the LLM~\cite{houlsby2019parameter}, or swap state for a significant number of parameters during inference.

A novel characteristic of our contribution to the PET field is that we explore the trade-offs when one has more data than can fit into the LLM's input context for ICL, but less than is effective for fine-tuning: in the range of tens of examples to 2,000. We argue that this is an important scale of data to consider as it represents what a small organization is likely to meaningfully create for a specialized text classification. 

\section{Datasets}

Our datasets cover three broad domains related to the quality of online discourse: classification of offensiveness to support safer chatbot dialogue   (Section~\ref{sec:dialog-safety}), attributes of online comments related to toxicity that result in unhealthy conversations (Section~\ref{sec:ucc}), and expressing responses in a neutral way, inspired by Wikipedia's policies (Section~\ref{sec:neutral-answers}).

\subsection{Dialogue safety}
For the dialogue safety domain, we consider 5 datasets for our experiments based on the \parlai{}~\cite{dinan-etal-2019-build} and Bot Adversarial Dialogue~\cite[BAD;][]{xu2021bot} data collection efforts.

\paragraph{\parlai{}.} We consider three independent datasets from \parlai{}, namely \parlaistandard{}, \parlaiadversarial{}, and \parlaimulti{}. All three datasets come with a pre-defined split as introduced by~\citet{dinan-etal-2019-build}, containing 24,000 training examples, and 3,000 each for validation and testing. 

\parlaistandard{} and \parlaiadversarial{} are both single-turn conversational datasets. For the former, crowdworkers were simply asked to construct sentences that they would consider offensive. The latter, in contrast, was built by asking crowdworkers to submit sentences that are \textit{offensive}, but are predicted to be \textit{safe} by a classifier. 

\parlaimulti{} is a multi-turn conversational dataset, consisting of sequences representing multiple turns of human conversations in which the last utterance is meant to be offensive. 

\paragraph{Bot Adversarial Dialogue.} The Bot Adversarial Dialogue (BAD) dataset, presented in~\citet{xu2021bot}, contains a collection of dialogues related to conversational toxicity. The dataset was collected by asking humans to converse with a bot, with the intention to lead the bot into generating \textit{offensive} output. The dataset comes with a predefined split of 5,080 conversations for training, 513 for validation, and 191 for testing. 

Since the conversations in the dataset can be lengthy (up to 14 turns), we follow~\citet{xu2021bot} and truncate conversations. Specifically, we experiment with versions of the dataset where we only consider the last four (\badfour{}) and two (\badtwo{}) utterances of the conversation.

\subsection{Unhealthy Comment Corpus}
The \textit{Unhealthy Comment Corpus}~\cite[UCC;][]{price-etal-2020-six} consists of 44,355 comments from the Globe and Mail news site (randomly sampled from the Simon Fraser University Opinion and Comment Corpus dataset~\cite{kolhatkar-2020}), labeled by crowdworkers for 7 labels.\footnote{For our experiments, we use the version of UCC available at \url{https://github.com/conversationai/unhealthy-conversations}.} Each comment is labeled as either \textit{healthy} or \textit{unhealthy}, in addition to binary labels for the presence of six unhealthy sub-attributes: (1) \textit{hostile}; (2) \textit{antagonistic}, \textit{insulting}, \textit{provocative} or \textit{trolling}; (3) \textit{dismissive}; (4) \textit{condescending}; (5) \textit{sarcastic}; and (6) \textit{generalization}.
It is worth mentioning that this dataset is highly imbalanced, with positive examples (the unhealthy attributes) comprising less than 10\% across splits and attributes (details can be found in Table~\ref{table:UCC_dataset_distribution} in the Appendix). This is typical of datasets sourced from online discourse labeled with forms of toxicity.

\subsection{Neutral Responses}

\begin{table}[!ht]
    \centering
    \resizebox{\columnwidth}{!}{
    \begin{tabular}{l c c}
    \toprule
     \textbf{Attribute} & \textbf{Total (pos. percentage)} & \textbf{Human AUC} \\
     \midrule
     Multiple perspectives & 113 (75.33 \%) & 0.965 $\pm$ 0.023\\
     Neutral & 76 (50.67\%) & 0.960 $\pm$ 0.025\\
     Well-explained & 91 (60.67\%) &  0.876 $\pm$ 0.040\\
     \bottomrule
    \end{tabular}}
    \caption{Statistics of the Neutral Responses dataset, including the absolute number as well as percentage of positive examples per attribute, as well as the Human AUC baseline. Each example was labeled by three expert annotators.}
    \label{table:NPOV_dataset_distribution}
\end{table}

To experiment with the effectiveness of prompt-tuning on a novel task, we built a new  Neutral Responses dataset comprised of human-written texts annotated according to three attributes that one might want from a chatbot's response to a difficult or polarizing question. The dataset contains 150 examples, each composed of a topic, a question regarding the topic, an answer to the question, and \textit{yes}/\textit{no} annotations labeling the answer along three attributes: whether it covers multiple perspectives on the topic, whether it is written from a neutral perspective (without expression opinion or judgement), and whether it is well-explained (see Appendix~\ref{app:linguistic_neutrality} for additional dataset details and examples).
The 150 examples span 75 different topics which were selected to be controversial. Each question and answer was written by an expert annotator.\footnote{The questions and answers on each topic can be opinionated and do not reflect the views of the authors.} This expert and two more experts then annotated the data along each attribute. The final label is obtained by majority vote. 

The distribution of attributes is shown in Table~\ref{table:NPOV_dataset_distribution}. The answers were crafted to limit the class imbalance so there are sufficiently many positive and negative examples to train a classifier. Additionally, each consistent triplet of values of attributes (multiple perspectives, \textit{not} neutral, \textit{not} well-explained) is also represented in a relatively balanced proportion: every triplet of values comprises between 12\% and 34\% of the dataset.\footnote{Note that some attribute labels are incompatible.} In our experiments, when sampling the splits for train and test sets, we use stratified sampling to approximately match the label distribution of the overall training set.

The overall inter-annotator agreement measures for both Krippendorff's alpha and Fleiss's kappa are 0.72. This is a strong agreement for subjective tasks of this nature; for reference, inter-annotator agreement for crowdsourced attributes related to online toxicity typically falls into the 0.4--0.6 range~\cite{wulczyn2017ex}. We also compute these measures by attributes: the \textit{Multiple perspectives} attribute achieves a Krippendorff's alpha of 0.76 (resp. Fleiss's kappa of 0.76), \textit{Neutral} achieves 0.79 (resp. 0.80), and \textit{Well-Explained} 0.59 (resp. 0.60). 

To provide a strong baseline for classifier quality for each attribute, we also compute the AUC for each annotator against the majority vote, and report the average as the human mean AUC; this is artificially high as the annotator contributes to the majority baseline. We also see that as the mean AUC for these tasks decreases, the standard deviation increases. This is an aspect of agreement measured by AUC, and correlates with subjective feedback from the annotators on task difficulty.

\section{Models and prompt-tuning}

For all three domains, we train soft prompts based on \tfivexxl{}~\cite{raffel2020exploring} and \palm{}~\cite{chowdhery2022palm}. Soft prompt tokens have embedding dimensionality according to the language model, which results in 4,096- and 8,192-dimensional embeddings for \tfivexxl{} and \palm{}, respectively. We follow~\citet{lester-etal-2021-power} and use an adapted version of \tfivexxl{}, which has been trained on a prefix language modeling objective. 
For all prompt-tuning experiments we follow~\citet{lester-etal-2021-power} and initialize each prompt with a random sample of vocabulary token embeddings from the respective model's 5,000 most frequent tokens.

We train soft prompts on 10, 20, 50, 80, 100, 200, 500, 1,000, and 2,000 randomly sampled training examples. For the Neutral Responses dataset, we use only up to 100 examples due to the dataset's more limited size. To account for variability in the sampling process, we repeat each experiment three times with different seeds and report average scores.

\paragraph{Dialogue safety and UCC.} Soft prompts consist of 10 tokens, resulting in a total of 40,960 (\tfivexxl{}) and 81,920 (\palm{}) learnable parameters per task, constituting only a small fraction of the 11 and 62 billion total parameters of each LLM. We train each prompt for 20 epochs, validate the loss after each epoch on a sampled subset of 500 validation examples, and select the best-performing checkpoint for testing. Each experiment uses the Adam optimizer~\cite{kingma2014adam} with a learning rate of 0.1 and weight decay 0.00001. For \palm{} we use a batch size of 4, and for \tfivexxl{} one of 32. 

\paragraph{Neutral Responses dataset.} Soft prompts of 5 tokens are trained for 1,000 steps for both \palm{} and \tfivexxl{}. We use the Adafactor optimizer~\cite{shazeer2018adafactor} with a constant learning rate of 0.1, a batch size of 16 and a prompt of length 5. For \tfivexxl{}, we also set the dropout rate to 0.1.

\paragraph{Evaluation metrics.} For the dialogue safety experiments we report model performance in terms of binary $F_1$ on the positive (toxic) class, in line with experiments in~\citet{dinan-etal-2019-build}. For both the Unhealthy Comments and Neutral Responses datasets, we report ROC-AUC scores. We obtain classification scores in the zero to one range from the LLMs by scoring specific tokens corresponding to the output class labels (e.g., the tokens \textit{yes} and \textit{no}), applying softmax, and then taking the score value of the positive class. This provides a threshold-agnostic and dataset-imbalance-agnostic metric, and allows comparison to the previous reported performance results.

\section{Baselines}

\subsection{In-context learning}

We compare prompt-tuning to \emph{in-context learning} (ICL) baselines, which include training data directly in the prompt-template sent to the model. We conduct experiments with zero-shot, 6-shot and 12-shot prompt-templates. For the latter two, for each seed we sample a fixed set of few-shot examples to be used for the inference prompt. For both few-shot learning and prompt-tuning, half of the training set is sampled from the positive class, half from the negative class. We repeat the few-shot experiments three times with different seeds to account for variation due to random sampling.

\subsection{Perspective API}
For the dialogue safety and online comments tasks, we further compare prompt-tuning with the Perspective API baseline: an off-the-shelf toxicity classifier which computes a toxicity confidence value for a given input text.\footnote{\url{https://perspectiveapi.com/}} In our experiments, we consider the eight attributes \texttt{TOXICITY}, \texttt{SEVERE\_TOXICITY}, \texttt{IDENTITY\_ATTACK}, \texttt{INSULT}, \texttt{PROFANITY}, \texttt{THREAT}, \texttt{FLIRTATION} and \texttt{SEXUALLY\_EXPLICIT}. For each test set example, we compute the Perspective score individually for each category, and use a threshold-based approach with threshold values of 0.0, 0.1, 0.2, \dots, 0.9 to predict whether a given piece of text is toxic. In the results, we only report the highest achieved performance across thresholds and attributes.

\section{Results}
\begin{table*}[!ht]
    \centering
    \resizebox{\textwidth}{!}{
        \begin{tabular}{l c c c c c c c c}
        \toprule
        & \multicolumn{5}{c}{\textbf{Dialogue Safety}} & \multicolumn{3}{c}{\textbf{Neutral  Responses}} \\
        \cmidrule(lr){2-6} \cmidrule(lr){7-9}
        \textbf{Model} & \textbf{\shortstack{\textsc{ParlAI} \\ \textsc{Single} \\ \textsc{Standard}}} & \textbf{\shortstack{\textsc{ParlAI} \\ \textsc{Single} \\ \textsc{Adversarial}}}  & \textbf{\shortstack{\textsc{ParlAI} \\ \textsc{Multi}}} & 
        \textbf{\shortstack{\textsc{BAD-2}}} & 
        \textbf{\shortstack{\textsc{BAD-4}}} &  
        \textbf{\shortstack{Multiple \\ Perspectives}} & \textbf{Neutral} & \textbf{\shortstack{Well- \\ Explained}} \\ \midrule
        \palm{} best few-shot & 0.89 & 0.67& 0.56 & 0.54 & 0.54 & 0.84 & 0.87 & 0.87 \\
        \midrule
        \tfivexxl{} - 80 & 0.18 & 0.18 & 0.19 & 0.29 & 0.48 & 0.94 &  0.96 & 0.76 \\
        \tfivexxl{} - 2,000 & 0.90 & 0.91 & 0.48 & 0.20 & 0.44 & --- & --- & --- \\
        \midrule 
        Human Agreement & --- & --- & --- & --- & --- &  \bftab 0.94 & 0.95 & \bftab 0.90\\
        Previous SOTA  & 0.88 & 0.67 & 0.66 & --- & --- & --- & --- & --- \\
        \midrule 
        \palm{} - 80  & 0.87 & 0.77 & 0.71 & 0.60 & 0.65 & 0.94 & \bftab 0.96 & 0.88 \\
        \palm{} -   2,000  & \bftab 0.95 & \bftab 0.91 & \bftab 0.81 & \bftab 0.68 & \bftab 0.70 & --- & --- & ---\\
        \bottomrule
        \end{tabular}
    }
    \resizebox{\textwidth}{!}{
        \begin{tabular}{l c c c c c c c  }
        \toprule
        & \multicolumn{7}{c}{\textbf{Unhealthy Comment Corpus}} \\
        \cmidrule(lr){2-8}
        \textbf{Model} &
        \textbf{Antagonistic} & \textbf{Condescending} & \textbf{Dismissive} & \textbf{Generalization} & \textbf{Hostile} & 
        \textbf{Sarcastic} & \textbf{Unhealthy}  \\ \midrule
        \palm{} best few-shot & 0.79 & 0.78 & 0.81 & 0.76 & 0.79 & 0.76 & 0.70 \\
        \midrule
        \tfivexxl{} - 80 & 0.50 & 0.55 & 0.56 & 0.49 & 0.57 & 0.54 & 0.51 \\
        \tfivexxl{} - 2,000 & 0.74 & 0.74 & 0.75 & 0.80 & 0.80 & 0.74 & 0.66 \\
        \midrule 
        Human Agreement & 0.71 & 0.72 & 0.68 & 0.73 & 0.76 & 0.72 & 0.62 \\
        Previous SOTA  & 0.82 & 0.78 & 0.82 & 0.74 & 0.84 & 0.64 & 0.69 \\
        \midrule 
        \palm{} - 80  & 0.80 & 0.80 & 0.74 & 0.81 & 0.84 & 0.81 & 0.63 \\
        \palm{} - 2,000  & \bftab 0.86 & \bftab 0.84 & \bftab 0.87 & \bftab 0.90 & \bftab 0.89 & \bftab 0.85 & \bftab 0.77 \\
        \bottomrule
        \end{tabular}
    }
    \caption{Summary of results for the dialogue safety ($F_1$ score), Neutral Responses (ROC-AUC), and Unhealthy Comments datasets (ROC-AUC), averaged over three seeds. We compare ICL for \palm{} and \tfivexxl{} across 0, 6, and 12 shots on a validation set, and report the results of the best model (\palm{}, either 6 or 12 shots). We add the human agreement baseline and the previous state-of-the-art~\cite{dinan-etal-2019-build} for dialogue safety, and results with BERT for the Unhealthy Comment Corpus~\cite{price-etal-2020-six}. \tfivexxl{} prompt-tuned on 2,000 examples outperforms the human agreement. \palm{} prompt-tuned on 80 examples also outperforms human agreement, and achieves SOTA. \palm{} prompt-tuned on 2,000 examples shows that the quality of results keeps improving as the dataset size increases.} 
    \label{table:all_results}
\end{table*}

\subsection{Dialogue safety}
\label{sec:dialog-safety}

Performance results on the dialogue safety datasets can be found in Table~\ref{table:all_results}, while Figure~\ref{fig:prompt_tuning_intro_figure} shows the quality of prompt-tuning as the number of training examples increases on the \parlaiadversarial{} dataset. The plots for the \parlaistandard{}, \parlaimulti{}, \badtwo{} and \badfour{} datasets are similar, and can be found in Appendix~\ref{app:dialog_safety}. For each dataset, we show the best-performing few-shot baselines. For the three \parlai{} datasets we also show the test set scores reported in~\citet{dinan-etal-2019-build}, obtained by fine-tuning BERT-Base on the entire training set containing 24,000 samples in each respective dataset. 

We observe that across the dialogue safety datasets, prompt-tuning \palm{} outperforms \tfivexxl{} when trained on both 80 and 2,000 examples. Furthermore, we can see that for \tfivexxl{}, there is a critical change in behavior as the training data increases (e.g., 2,000 instead of 80): in two cases, \tfivexxl{} jumps from being significantly worse than random to outperforming the previous state-of-the-art (e.g., increasing from 0.18 to 0.91 $F_1$ on \parlaiadversarial{}).\footnote{Note that $F_1$ scores below 0.50, that of a random classifier, happen when the data is imbalanced, and for smaller datasets \tfivexxl{} biases towards predicting the majority class (which leads to high accuracy, but low $F_1$ as recall is low on the minority class).} Such differences are much less clear for \palm{}, since the model performs well with as little as 80 examples across datasets, where it already outperforms the previous state-of-the-art~\cite{dinan-etal-2019-build}.

Taking a closer look at the differences between few-shot learning and prompt-tuning, we observe that utilizing \palm{} for few-shot learning also suffices to perform on par with  (\parlaiadversarial{}) or outperform (\parlaistandard{}) the previous state-of-the-art, indicating that few-shot learning represents a competitive baseline in this setting. However, prompt-tuning on 80 training examples suffices to outperform the few-shot ICL in four out of the five tasks. Note that a comparison to few-shot learning with 80 examples is not possible due to the context window restrictions for \tfivexxl{} and \palm{}.

Both LLMs perform substantially better on single-turn datasets (\parlaistandard{}, \parlaiadversarial{}) compared to multi-turn datasets (\parlaimulti{}, \badtwo{}, \badfour{}). This demonstrates the difficulty of detecting safety concerns in multi-turn conversations.

\subsection{Unhealthy Comment Corpus}
\label{sec:ucc}

\begin{figure}[!ht]
     \centering
     \includegraphics[width=\columnwidth]{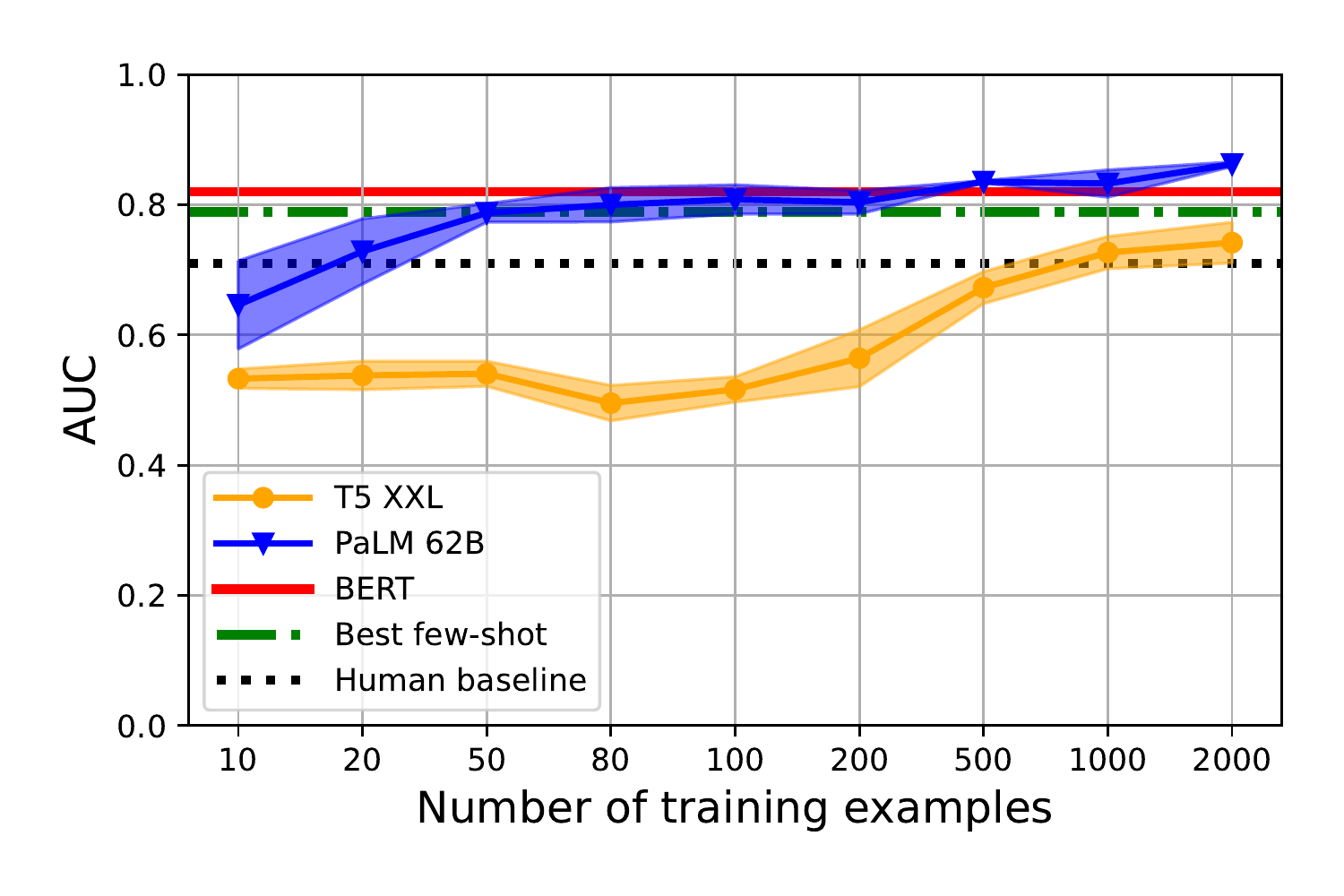}
    \caption{Prompt-tuning AUC results for the \textit{Antagonistic} attribute of the UCC dataset, across three seeds. \palm{} outperforms both 12-shot prompt on \palm{} and a BERT model fine-tuned on 44,335 training examples. The plot for the other attributes look similar and can be found in Appendix~\ref{app:prompt_tuning_ucc}.}
    \label{fig:prompt_tuning_antagonize}
\end{figure}

For UCC, like previous experiments, we ran few-shot and prompt-tuning experiments with \palm{} and \tfivexxl{} to classify comments for each class by prompting the models with a target text and the question \textit{Is the text above \{\texttt{class}\}?} This is inspired by the Question and Answering template presented in~\citet{DBLP:journals/corr/abs-2112-11446}, which adapts the verbalization methods proposed by~\citet{Schick2021SelfDiagnosisAS}.

To compare with the previous state-of-the-art, we report the performance in terms of ROC-AUC, and we use the same test set as \citet{price-etal-2020-six}. We contrast our results on the UCC dataset with their Human and BERT baselines, which evaluate performance according to assessments from human crowdworkers as well as a BERT model fine-tuned on the entire training dataset.

In Figure~\ref{fig:prompt_tuning_antagonize}, we show a typical example of how AUC varies as we use more training examples for prompt-tuning (results for the remaining attributes can be found in Appendix~\ref{app:prompt_tuning_ucc}). The trends are comparable to  Figure~\ref{fig:prompt_tuning_intro_figure}.

Quantitative results can be found in Table~\ref{table:all_results} (\textbf{Unhealthy Comment Corpus}). As can be seen, with only 80 training examples, prompt-tuning \palm{} outperforms the human baseline, and is comparable to the BERT baselines across all experiments. Performance scores for \palm{} increase further when training on more examples up to our upper bound of 2,000. However, as in the results discussed in Section~\ref{sec:dialog-safety}, the performance differences between 80 and 2,000 examples are much more drastic for \tfivexxl{}, showing absolute AUC improvements of around 0.3 for \textit{Generalization} and 0.2 for \textit{Antagonistic, Condescending, Dismissive, Hostile,} and \textit{Sarcastic}. Additionally, \tfivexxl{} outperforms the BERT baselines in two out of the five cases (\textit{Generalization} and \textit{Sarcastic}). While this shows that prompt-tuning \tfivexxl{} can be competitive with full model fine-tuning, it also demonstrates the benefit of using a larger model (i.e., \palm{}) for prompt-tuning on small datasets.

We also observe that the \palm{} few-shot baseline is competitive on many of the attributes in this dataset, nearly reaching or outperforming the BERT baseline in terms of AUC. However, few-shot ICL is also constrained by the context window that creates a hard limit on the number of examples that can be provided, and has high variance (depending significantly on the specific examples in the few-shot prompt). 

Overall, these results indicate that with fewer than 100 examples, prompt-tuning \palm{} achieves competitive performance across attributes, demonstrating its ability to serve as a method for efficiently building safety classifiers in data-scarce settings.

\subsection{Neutral Responses}
\label{sec:neutral-answers}
\begin{figure}[!ht]
         \centering
         \includegraphics[width=\columnwidth]{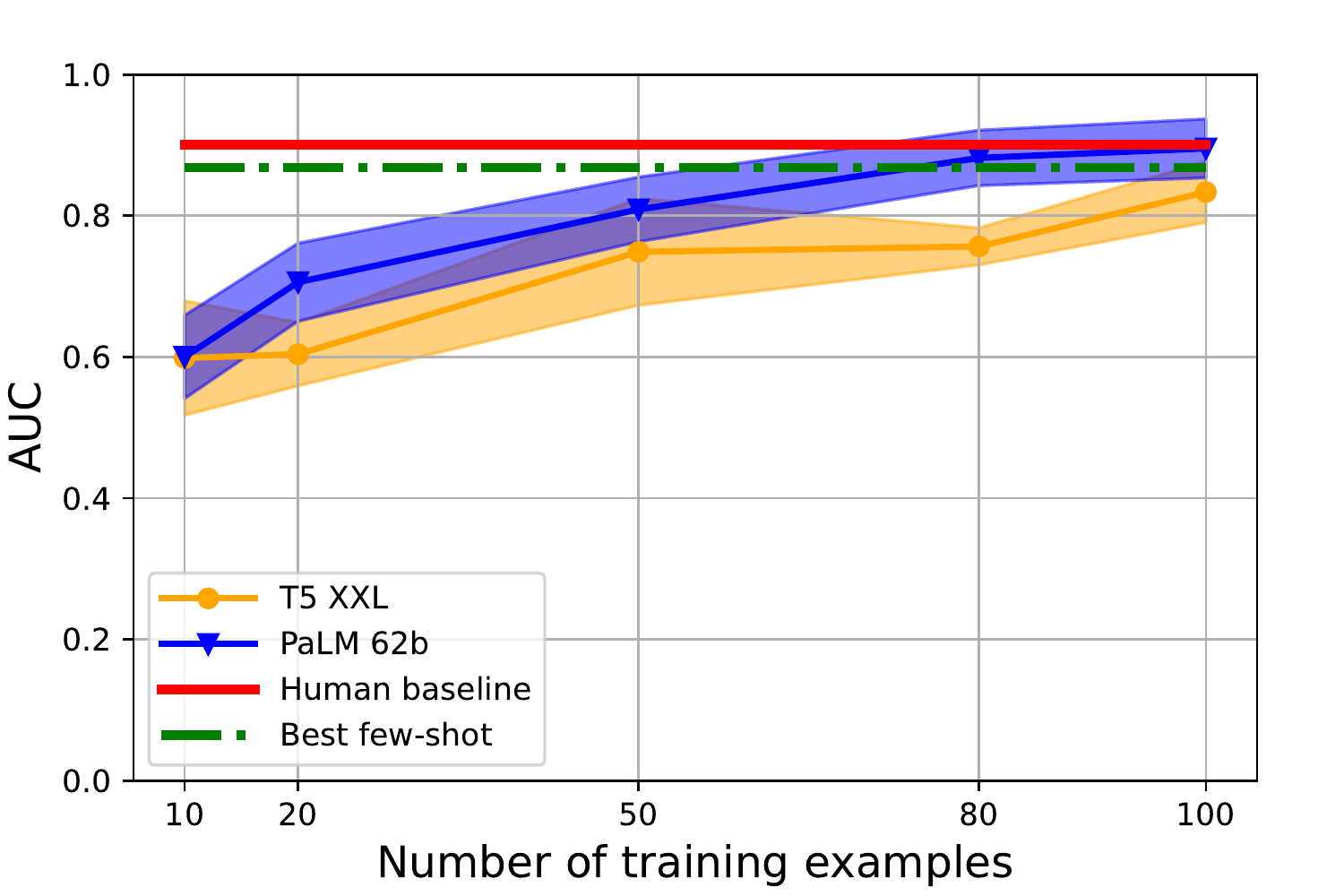}
     \hfill
    \caption{Prompt-tuning AUC results for the \textit{Well-Explained} attribute of the Neutral Responses dataset, across six seeds. \palm{} outperforms both 12-shot prompts and the human agreement baseline. Prompt-tuning exhibits similar behavior on the other attributes---results can be found in Appendix~\ref{app:prompt_tuning_linguistic_neutrality}.}
    \label{fig:prompt_tuning_NPOV}
\end{figure}

Quantitative results are reported in Table~\ref{table:all_results} (\textbf{Neutral  Responses}), and a typical plot is shown in Figure~\ref{fig:prompt_tuning_NPOV} (the plots for the remaining attributes can be found in Appendix~\ref{app:prompt_tuning_linguistic_neutrality}). From left to right, the tasks are shown by increasing order of complexity for humans. Unsurprisingly, the number of examples needed to achieve (or surpass) the human baseline increases with the complexity of the task. Like in our results with the other datasets, we observe that prompt-tuning with 80 examples is enough to get human-level accuracy (and to exceed few-shot prompting).

In contrast to the previous two domains, the difference between \tfivexxl{} and \palm{} appears to be smaller for the Neutral Responses dataset. This may be because it is an easier task, or because the training data quality is higher.

\section{Further Work}

We encourage researchers working towards healthier online discourse to further explore the new paradigm of small datasets and PET. While our results show that prompt-tuning is a strong method for our tasks, there are many other PET methods, e.g., prefix-tuning~\cite{liu2022few}, that may have different properties for text classification at the scale of data in between fine-tuning and ICL few-shot prompts.

The limits of PET for text classifiers are also important areas to explore. For example, one would expect poor results when prompt-tuning an LLM pretrained on one language in order to obtain an effective classifier in a different language. A deeper understanding of these limits will be important for the maturation of the field. In line with this, another interesting direction for future work could focus on the potential of instruction-tuned models for PET in this context, since such models have shown promising performances on unseen tasks~\cite{wei2021finetuned}.

Understanding unintended biases in small datasets will also become critical if PET becomes more widely adopted as a tool for agile classifier development. This is because datasets written by an individual may induce more unintended biases in the resulting classifiers since a smaller number of people review the policy. On the other hand, if the datasets are much smaller it may also become simpler to review and correct them in the underlying dataset.

Finally, another important branch of future work is to investigate augmenting, scaling, and replacing aspects of current human annotation with synthetic generation. Early work in this direction for Question and Answer tasks has recently proved successful~\cite{agrawal2022qameleon}. By combining synthetic generation with prompt-tuning, we speculate that a rich methodology can be created for the agile development of text classifiers.

\section{Conclusions}
In this paper, we demonstrated that we can use LLM-based parameter-efficient tuning to build high-performance classifiers with small datasets (e.g., of as few as 80 examples) across three domains: 5 datasets related to offensive dialogue (Section~\ref{sec:dialog-safety}), 7 annotation schemes related to toxicity in online comments (Section~\ref{sec:ucc}), and three attributes related to neutral responses to questions on sensitive topics (Section~\ref{sec:neutral-answers}).

We focused on the prompt-tuning approach to PET; it is one of the most parameter-efficient PET methods, allowing a single model to be served and the task-specialization to be provided in a soft prompt vector at query time, much like an ICL prompt-template. In contrast, fine-tuning requires changing many more model parameters and serving a separate model per task.

Our results show that prompt-tuning on small datasets consistently achieves performance that is competitive with the previous state-of-the-art (e.g., BERT-based fine-tuning approaches that use much larger datasets of human-annotated examples). Prompt-tuning performance also appears to be equal to or better than human-annotation quality. We found that ICL with few-shot templates is sometimes very effective, but also has much more variable performance. 

When prompt-tuning \tfivexxl{}, we observed that much more data is needed for effective performance on most datasets; the model only becomes competitive with fine-tuned baselines when trained on thousands of examples. In contrast, prompt-tuning on the much larger \palm{} model requires less than a hundred examples. This suggests that there is a form of scaling law for prompt-tuning, and that its utility dramatically improves as the model size grows.

We remark that the scale of data needed for prompt-tuning to achieve useful and even state-of-the-art performance in many tasks is now sufficiently small that a single individual could create the needed dataset. We argue that this represents a paradigm shift for text classification, especially in the domain of online safety: ICL and prompt-tuning large LLMs could enable even small forums to develop customized classifiers for their own evolving policies, rather than depend on centralized classifiers, trained on millions of annotations representing a single common policy. 

\bibliography{anthology,references}
\bibliographystyle{acl_natbib}

\appendix

\begin{table}[!ht]
    \centering
    \resizebox{\columnwidth}{!}{
        \begin{tabular}{l c c c}
        \toprule
        \textbf{Attribute} & \textbf{ Train} & \textbf{Test} & \textbf{Val}  \\
        \midrule
        Antagonistic & 1689 (4.8\%) & 203 (4.6\%) & 174 (3.9\%) \\
        Condescending & 1927 (5.4\%) & 269 (6.1\%) & 238 (5.4\%) \\
        Dismissive &  1071 (3.0\%) & 150 (3.4\%) & 143 (3.2\%) \\
        Generalization & 752 (2.1\%) & 96 (2.2\%) & 96 (2.2\%) \\
        Hostile & 923 (2.6\%) & 108 (2.4\%) & 99 (2.2\%) \\
        Sarcastic & 1501 (4.2\%) & 201 (4.5\%) & 195 (4.4\%) \\
        Unhealthy &  2655 (7.5\%) & 320 (7.2\%) & 366 (8.3\%) \\
        \bottomrule
        \end{tabular}
        }
    \caption{The absolute number as well as percentage of positive examples per split and attribute of the UCC dataset splits.}
    \label{table:UCC_dataset_distribution}
\end{table}

\section{Neutral Responses Dataset}
\label{app:linguistic_neutrality}

\begin{table*}[!ht]
    \centering
    \begin{tabular}{ | p{0.15\linewidth} | p{0.5\linewidth} | p{0.1\linewidth} | p{0.06\linewidth} | p{0.07\linewidth} |}
    \hline
     {\small Question} & {\small Answer} & {\small Multiple perspectives} & {\small Neutral} & {\small Well-explained} \\ 
     \hline
     {\footnotesize Everyone believes we should legalize marijuana, right?} &	{\footnotesize Some people believe legalizing marijuana would reduce crime and create jobs, while others oppose legalization because they believe canabis is a gateway drug and legalizing it will increase hard drug use.} & Yes & Yes & Yes \\
     \hline
     {\small What is this about taxes and churches?} & {\footnotesize Churches are currently tax exempt according to US law. There is a part of the population who wants to keep the status quo, and another who is awaiting a change.} & Yes & Yes & No \\
    \hline
    {\small Can you tell me about the water ban?}	& {\small I'm so tired of hearing about the water ban. Nobody cares.} & No	& No & No \\
    \hline
    \end{tabular}
    \caption{Three examples from the neutral responses dataset.}
    \label{table:examples_NPOV}
\end{table*}

\subsection{Detailed generative process of the dataset}
 The first expert gathered sources on each controversial topic, and was sometimes aided by two generative models: at first, they used a PaLM 540B model~\cite{chowdhery2022palm} using examples from the subreddit \textit{r/eli5} as few-shot. Once a small subset of the dataset was created, they used that subset as few-shot for a FLAN-PaLMChilla 62B model~\cite{chung2022scaling}. Most answers were written without that aid. Each answer written with that aid was heavily edited.
 
\subsection{Examples}
Examples from the neutral responses dataset can be seen in in Table \ref{table:examples_NPOV}.

\section{Prompt-tuning results for dialogue safety}
\label{app:dialog_safety}

\begin{figure*}[!ht]
     \centering
     \begin{subfigure}[b]{0.48\textwidth}
         \centering
         \includegraphics[width=\textwidth]{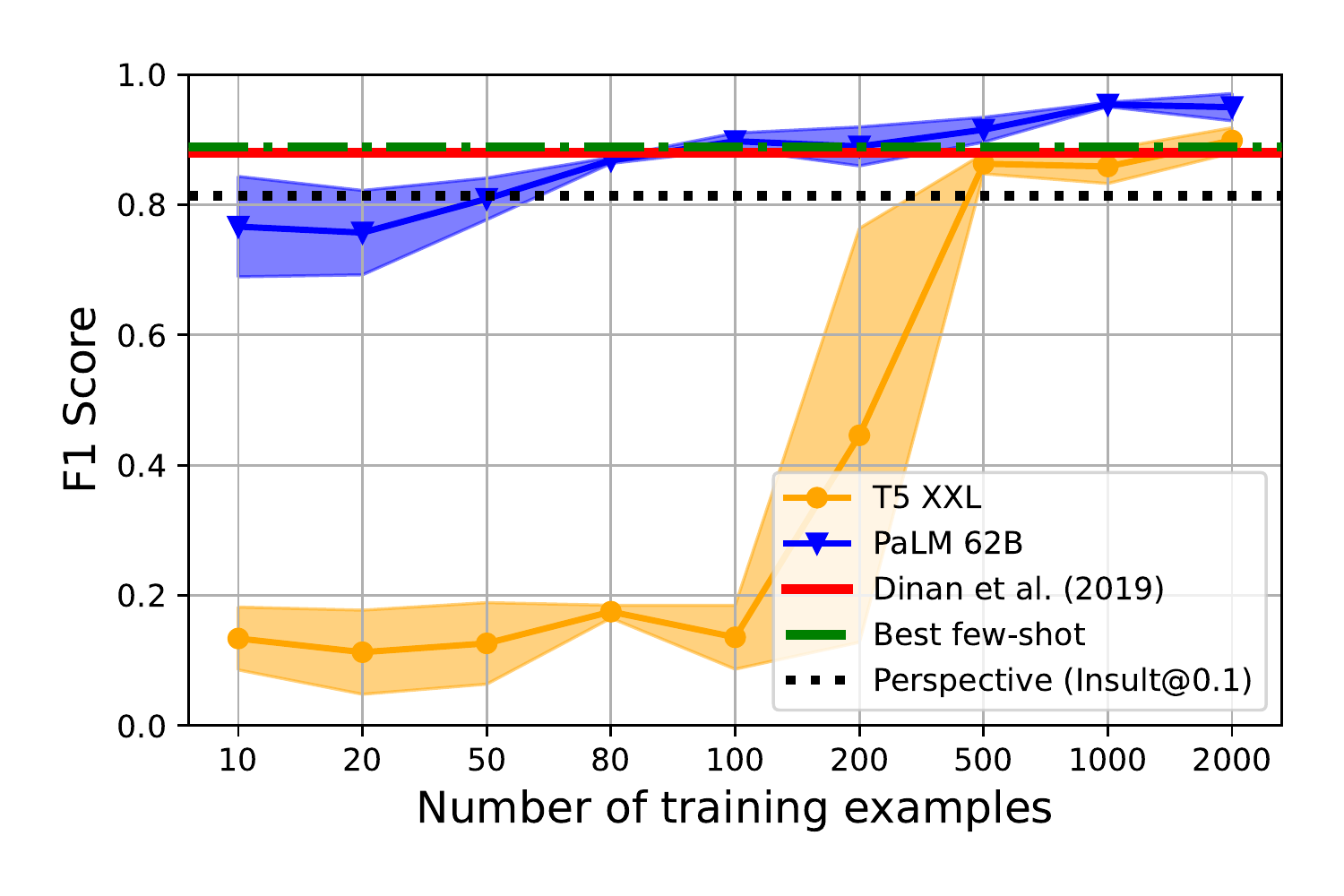}
         \caption{\parlaistandard{}}
     \end{subfigure}
     \hfill
     \begin{subfigure}[b]{0.48\textwidth}
         \centering
         \includegraphics[width=\textwidth]{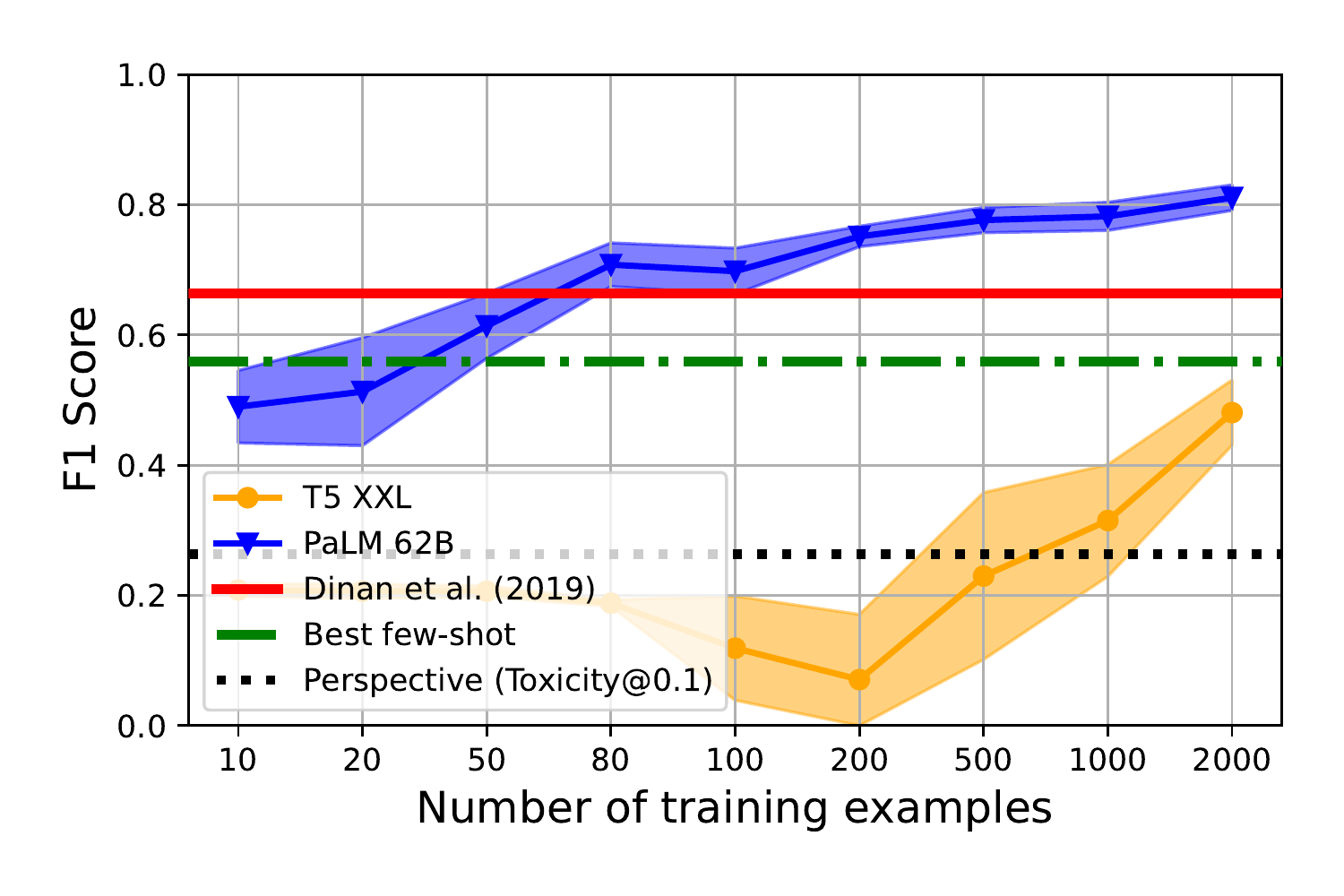}
         \caption{\parlaimulti{}}
     \end{subfigure}
     \hfill
     \begin{subfigure}[b]{0.48\textwidth}
         \centering
         \includegraphics[width=\textwidth]{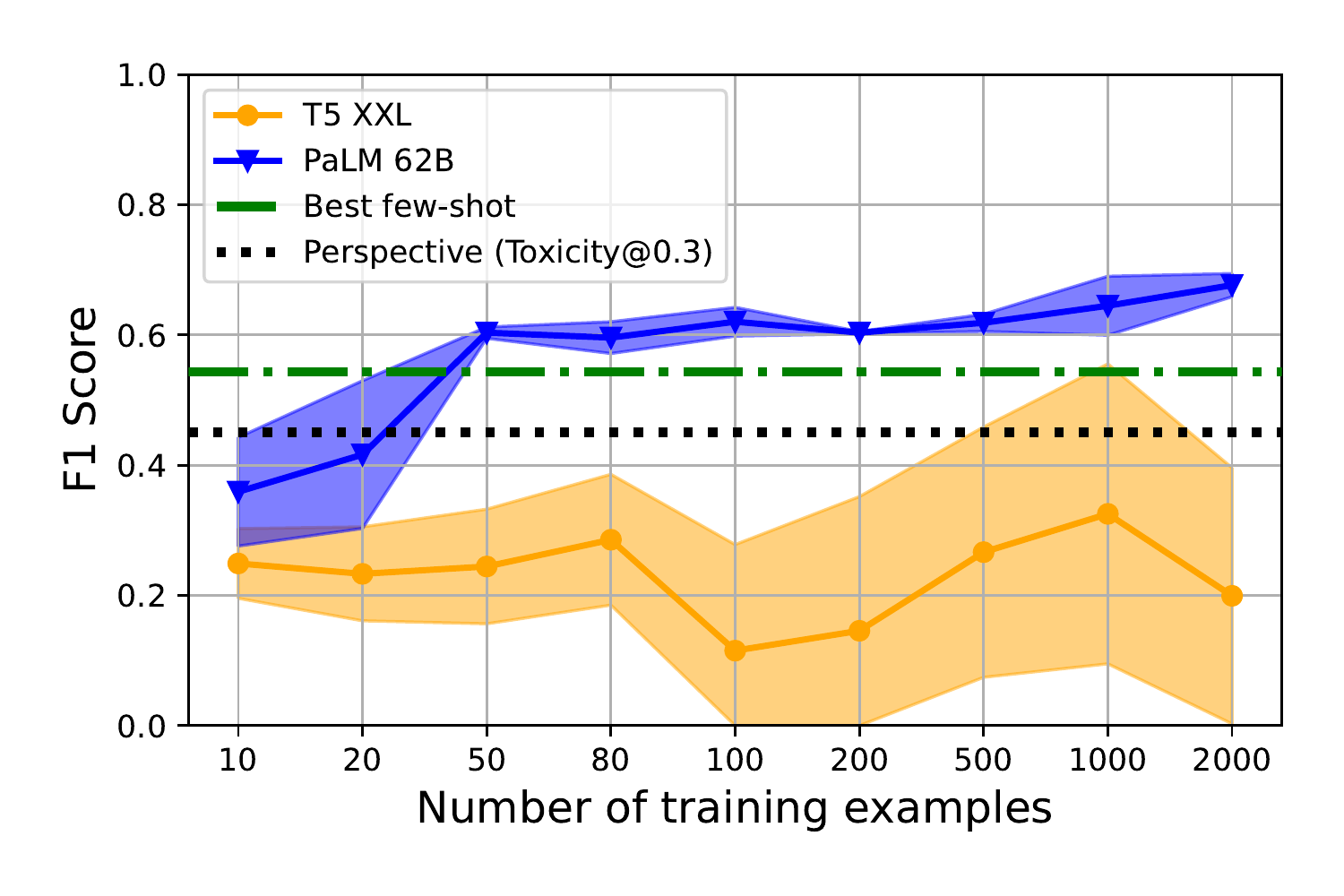}
         \caption{\badtwo{}}
     \end{subfigure} 
     \begin{subfigure}[b]{0.48\textwidth}
         \centering
         \includegraphics[width=\textwidth]{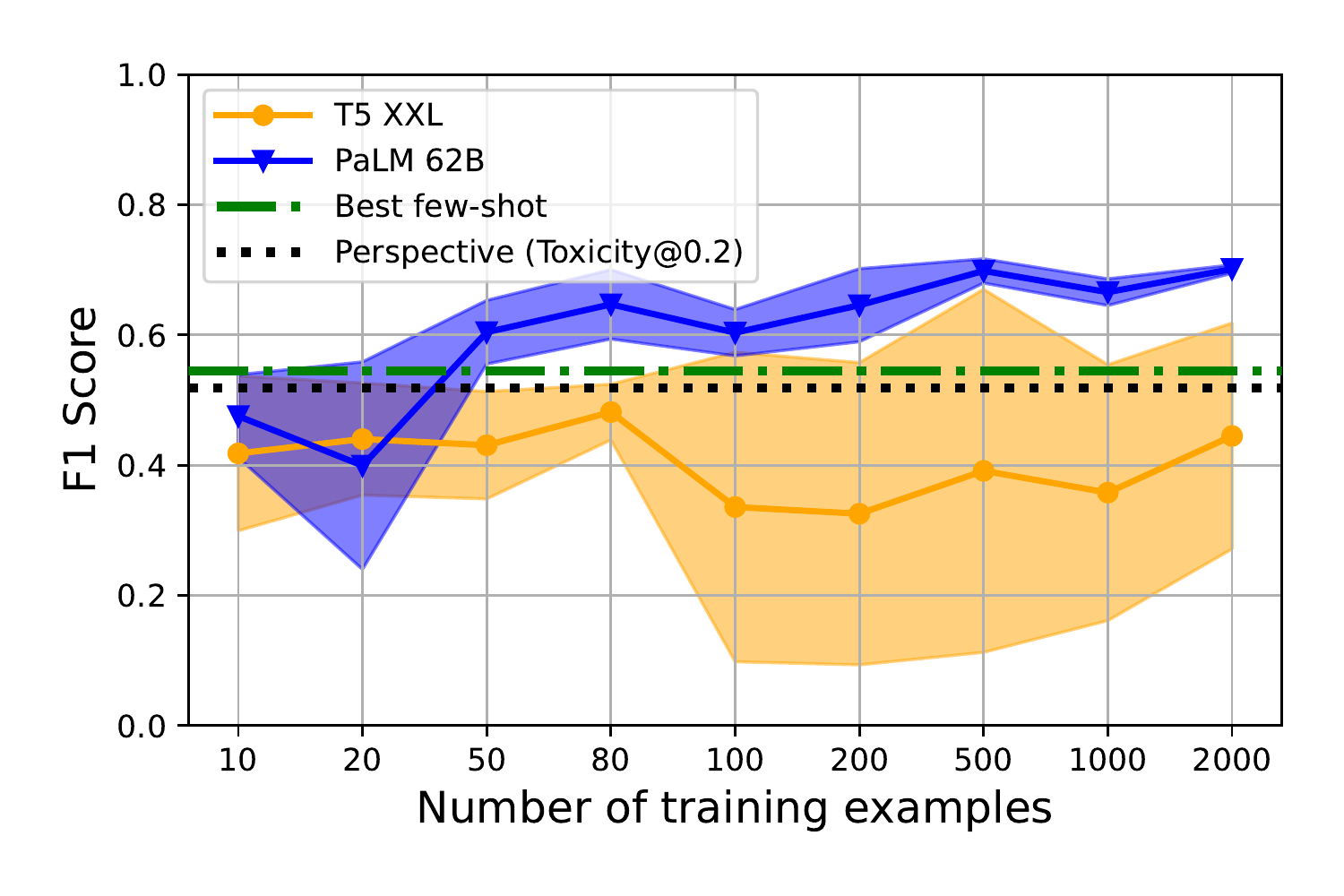}
         \caption{\badfour{}}
     \end{subfigure}
     \hfill
    \caption{Prompt-tuning results (test set $F_1$ averaged across three seeds with standard deviations visualized) for the \parlaistandard{}, \parlaimulti{}, and the two BAD datasets across models (\tfivexxl{}, \palm{}). Prompt-tuning performance is compared to few-shot learning and the Perspective API baseline (only the best scores across categories and thresholds are shown). For few-shot, we show \palm{}'s best performing score across 0, 6, and 12 shots.}
    \label{fig:prompt_tuning_dialog_safety}
\end{figure*}

Prompt-tuning results on the \parlaistandard{}, \parlaimulti{}, \badtwo{}, and \badfour{} datasets can be found in Figure~\ref{fig:prompt_tuning_dialog_safety}. In line with the results reported in Table~\ref{table:all_results}, we observe that prompt-tuning \palm{} with 100 (\parlaistandard{}) and 80 (\parlaimulti{}) examples suffices to outperform the performance of the fine-tuned BERT as reported in~\citet{dinan-etal-2019-build}. Similarly, for both \badtwo{} and \badfour{}, we can see that \palm{} outperforms both the Perspective and few-shot baselines when trained on 50 examples.

In contrast, \tfivexxl{} requires a larger amount of training data for prompt-tuning (2,000 examples) on \parlaistandard{} to outperform the BERT baseline. On the remaining three datasets, prompt-tuning \tfivexxl{} does not achieve a performance better than the few-shot baseline. 

\section{Prompt-tuning results for UCC}
\label{app:prompt_tuning_ucc}
\begin{figure*}[!ht]
     \centering
     \begin{subfigure}[b]{0.48\textwidth}
         \centering
         \includegraphics[width=\textwidth]{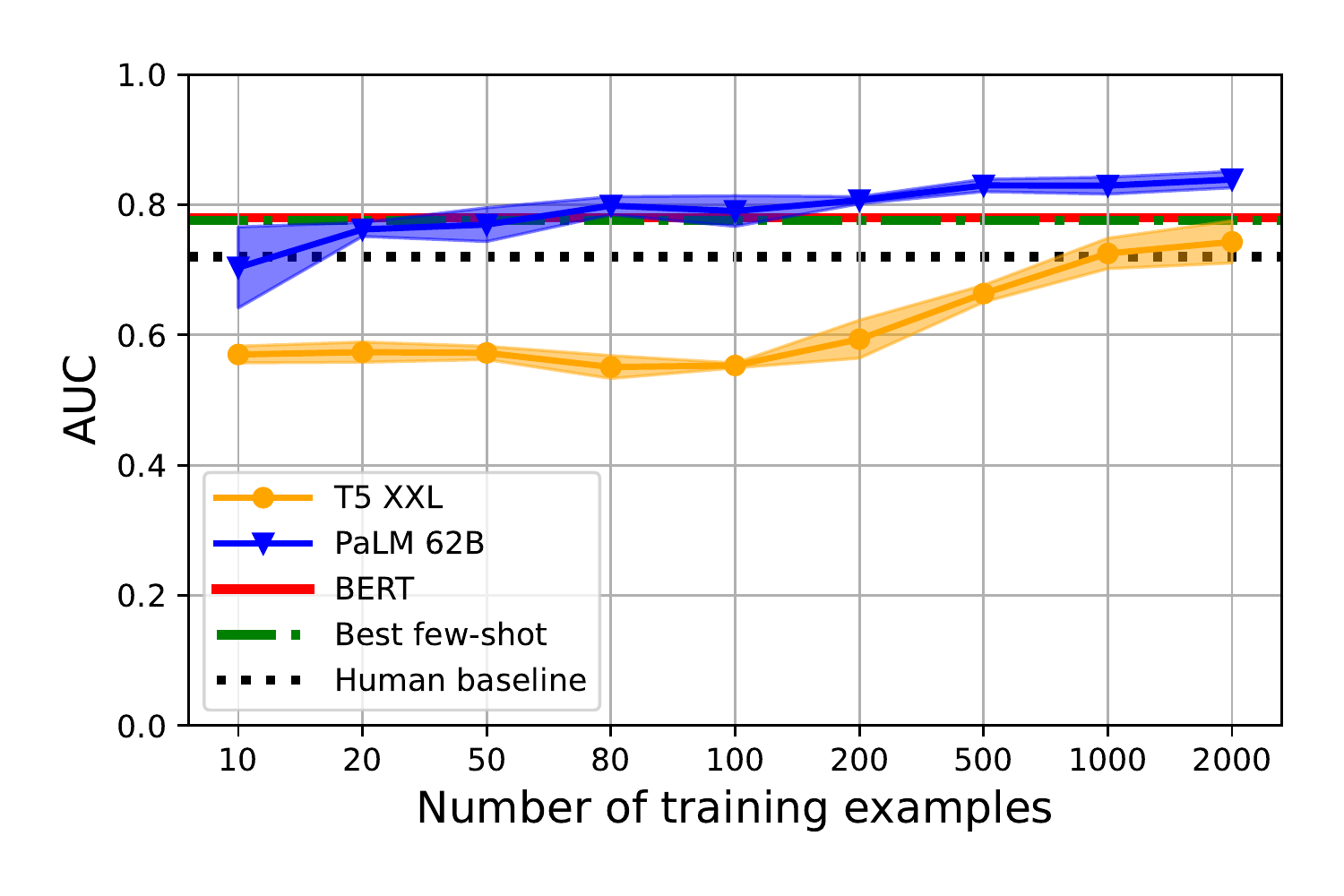}
         \caption{Condescending}
     \end{subfigure}
     \hfill
     \begin{subfigure}[b]{0.48\textwidth}
         \centering
         \includegraphics[width=\textwidth]{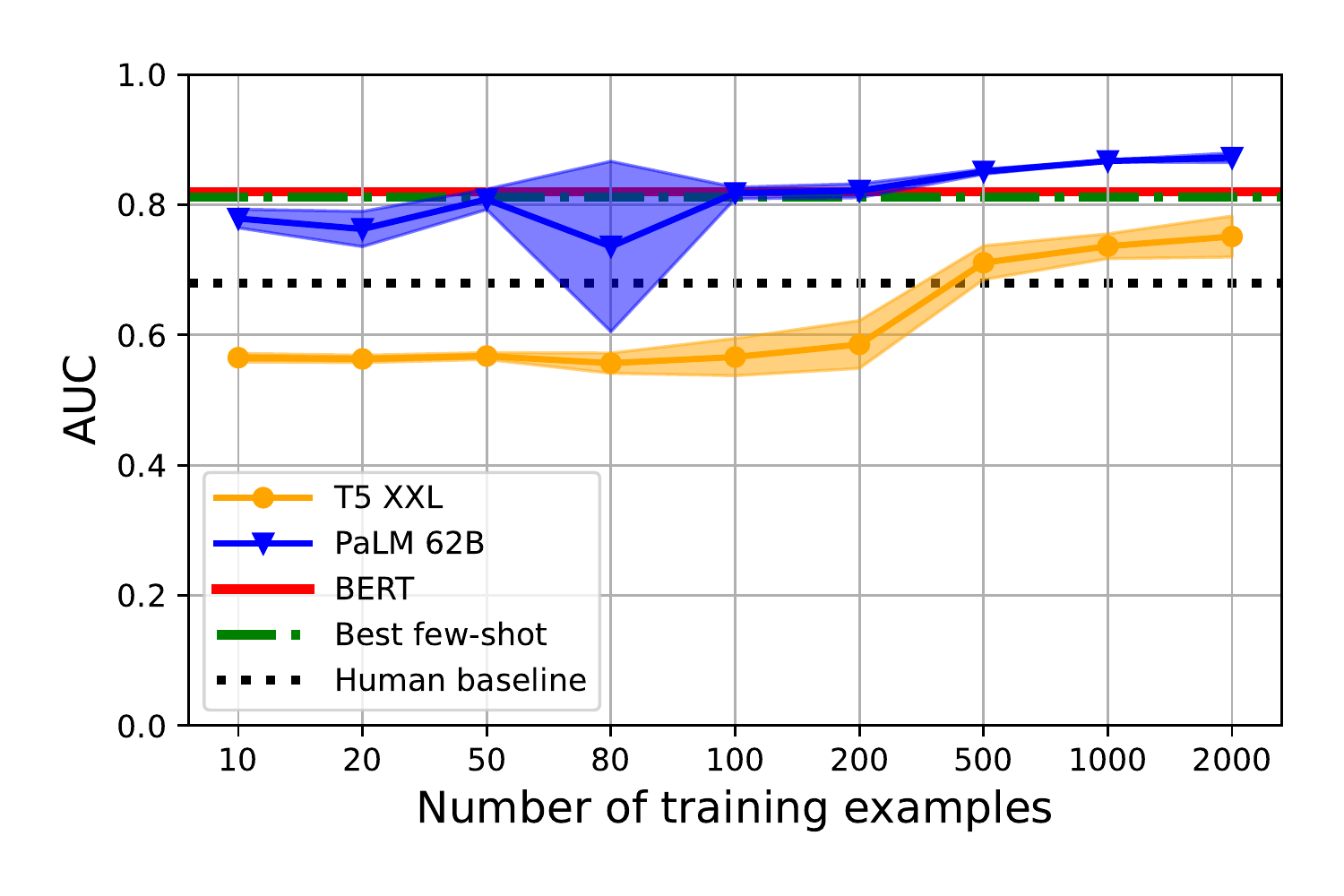}
         \caption{Dismissive}
     \end{subfigure}
     \hfill
     \begin{subfigure}[b]{0.48\textwidth}
         \centering
         \includegraphics[width=\textwidth]{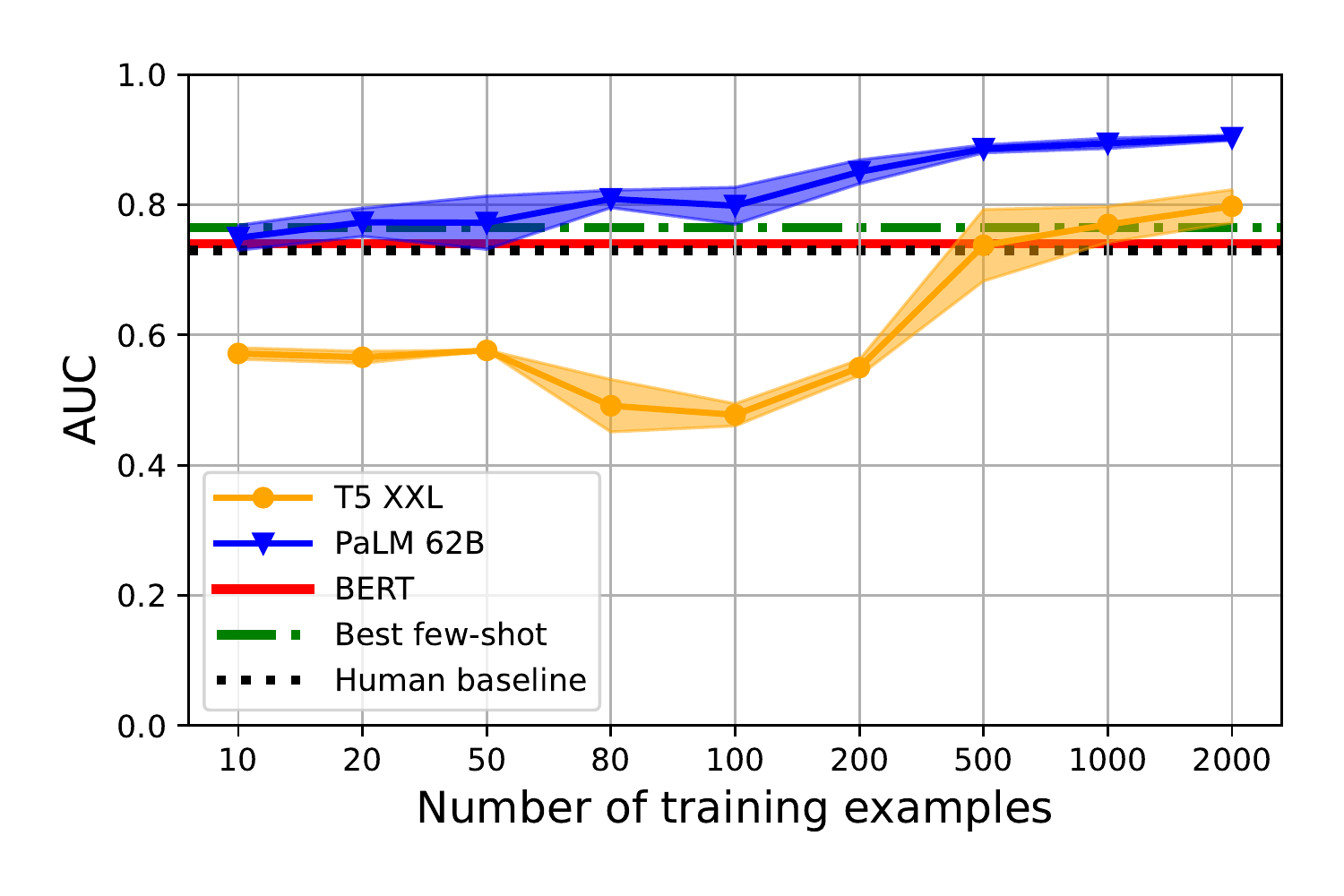}
         \caption{Generalization}
     \end{subfigure}
     \hfill
     \begin{subfigure}[b]{0.48\textwidth}
         \centering
         \includegraphics[width=\textwidth]{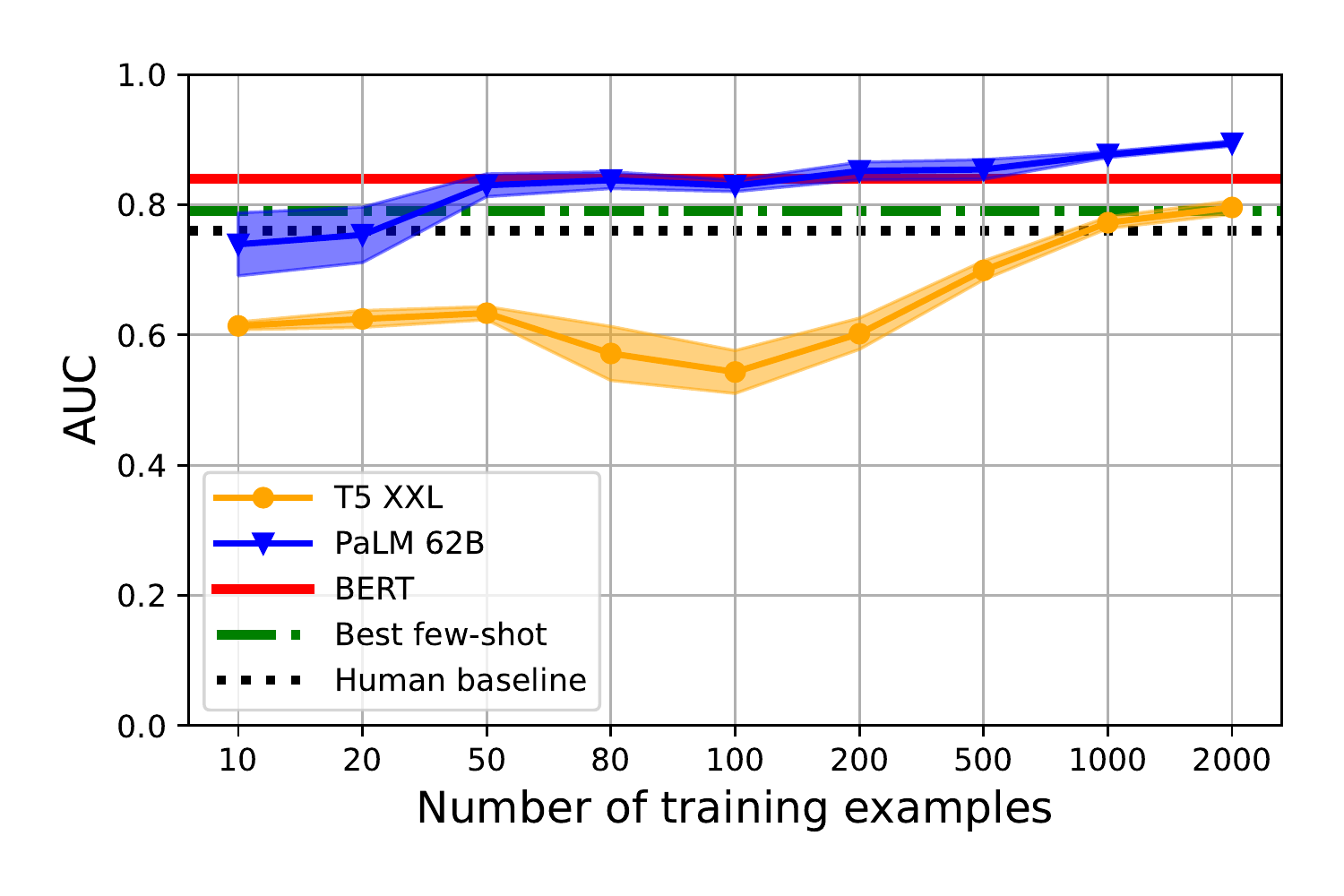}
         \caption{Hostile}
     \end{subfigure}
     \hfill
     \begin{subfigure}[b]{0.48\textwidth}
         \centering
         \includegraphics[width=\textwidth]{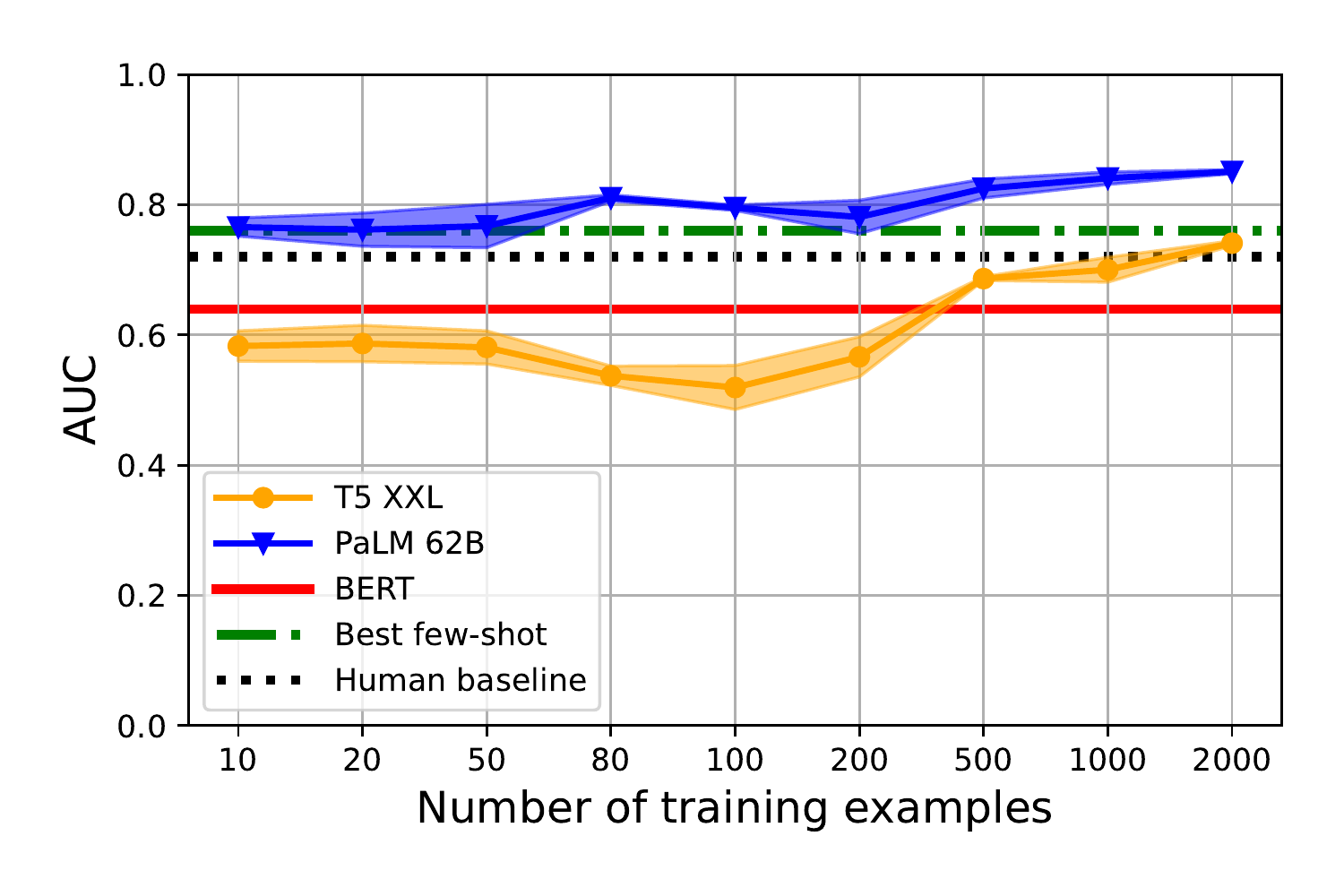}
         \caption{Sarcastic}
     \end{subfigure}
     \hfill
     \begin{subfigure}[b]{0.48\textwidth}
         \centering
         \includegraphics[width=\textwidth]{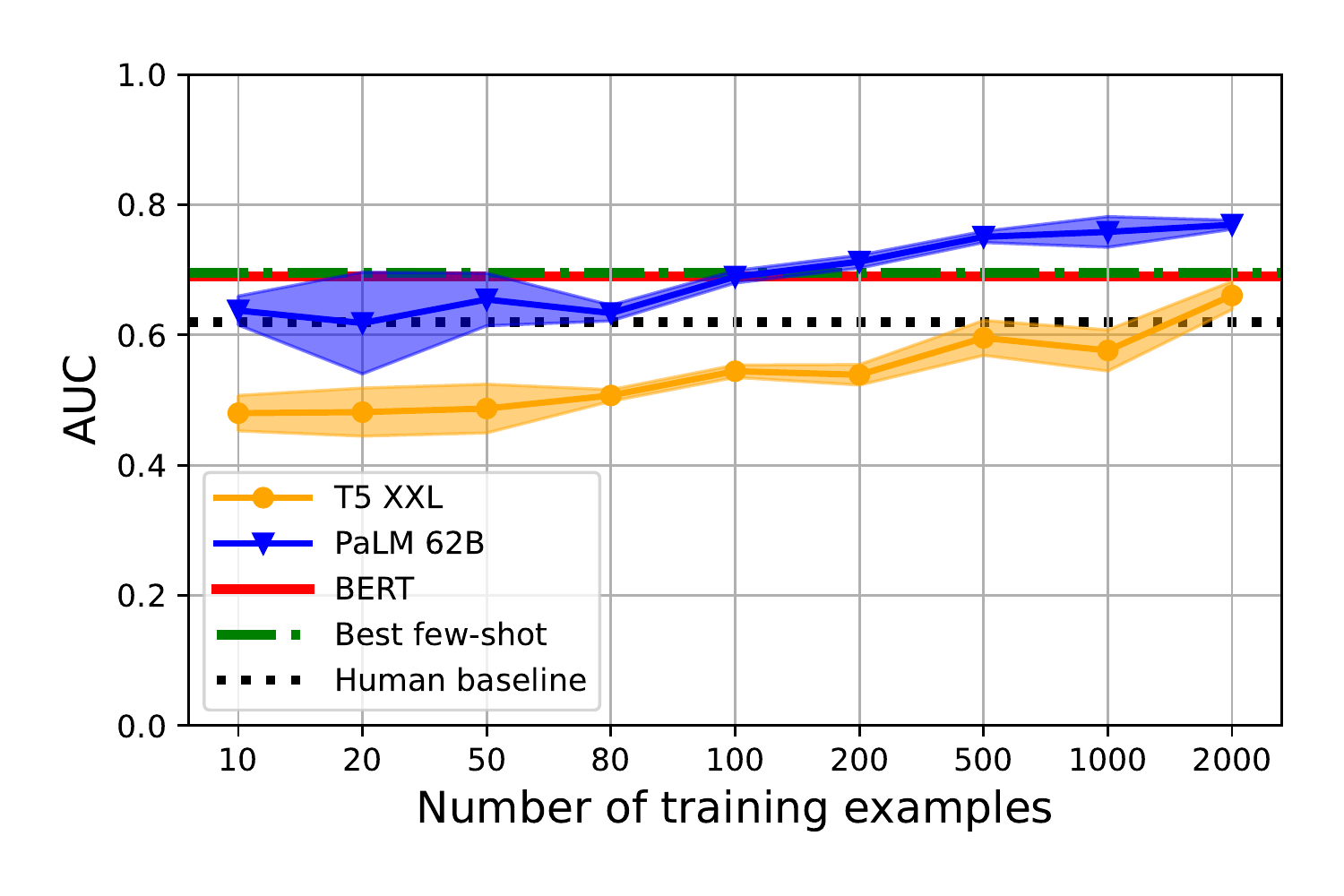}
         \caption{Unhealthy}
     \end{subfigure}
     \hfill
    \caption{Prompt-tuning results (ROC-AUC) on six attributes of the UCC dataset using \tfivexxl{} and \palm{}. \textsc{Human} and \textsc{BERT} denote the human and BERT-based baselines from~\citet{price-etal-2020-six}, respectively. For few-shot, we show \palm{}'s best performing score across 0, 6, and 12 shots.}
    \label{fig:prompt_tuning_ucc}
\end{figure*}

The prompt-tuning performances for the six remaining attributes of UCC can be found in Figure~\ref{fig:prompt_tuning_ucc}. Across attributes, we can see that prompt-tuning \palm{} achieves test set scores superior to the ones obtained with BERT fine-tuning. Interestingly, the model outperforms BERT with as little as 10 training examples (\textit{Generalization} and \textit{Sarcastic}). With 500 training examples, \palm{} outperforms BERT for all six attributes. 

The results for \tfivexxl{} reveal a different picture. While prompt-tuning achieves competitive performances across attributes, only for the attributes \textit{Generalization} and \textit{Sarcastic} does it manage to perform better than the BERT baseline. 

Taking a closer look at the few-shot baseline, we observe that it performs close to or outperforms BERT in five out of the six categories (only for \textit{Hostile} do we see a notable performance gap between the two). These results demonstrate the strength of the few-shot baseline, which is nevertheless outperformed by prompt-tuning \palm{} across all attributes.

\section{Prompt-tuning results for the Neutral Responses dataset}
\label{app:prompt_tuning_linguistic_neutrality}
\begin{figure*}[!ht]
     \centering
     \begin{subfigure}[b]{0.48\textwidth}
         \centering
         \includegraphics[width=\textwidth]{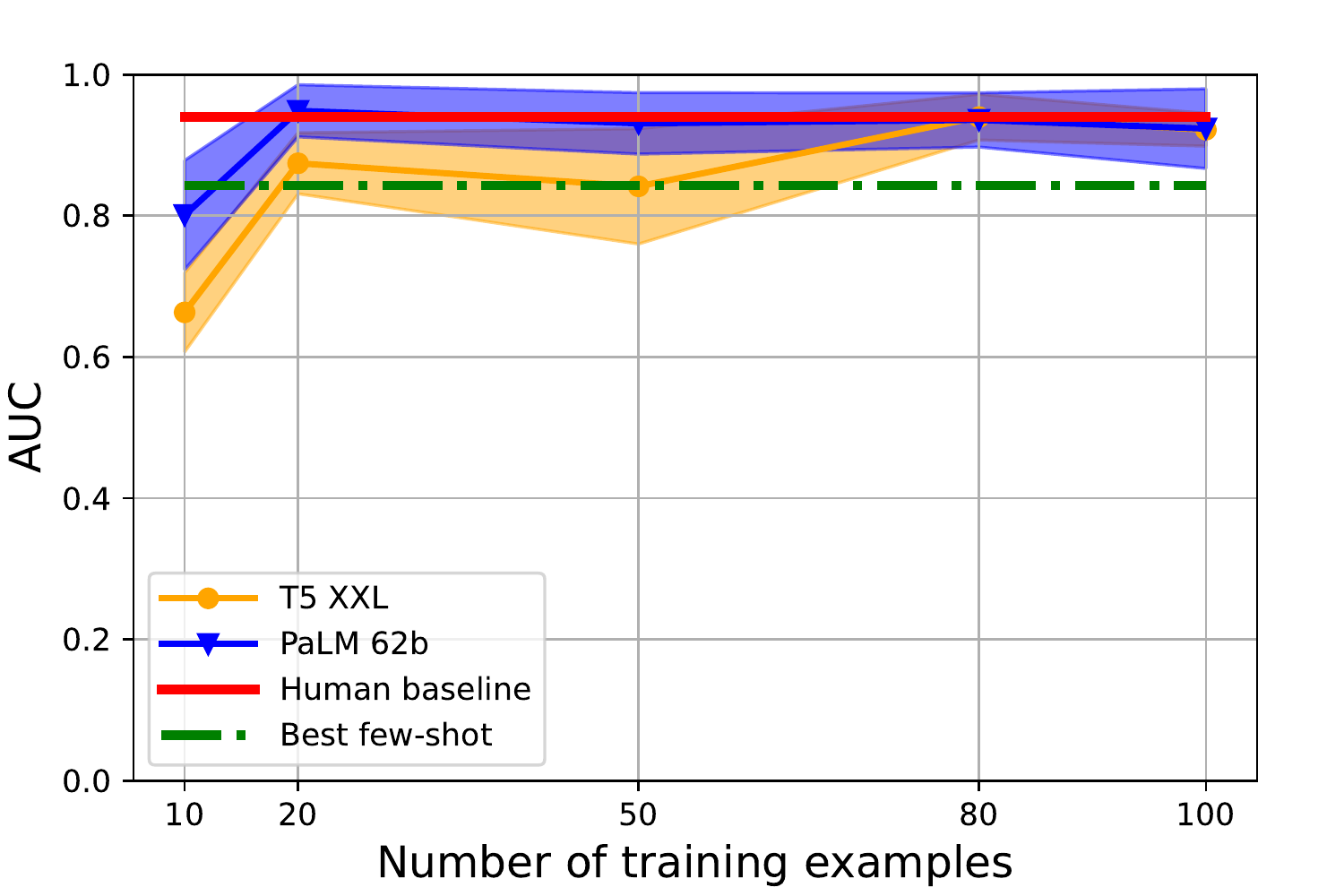}
         \caption{Multiple perspectives}
     \end{subfigure}
     \hfill
     \begin{subfigure}[b]{0.48\textwidth}
         \centering
         \includegraphics[width=\textwidth]{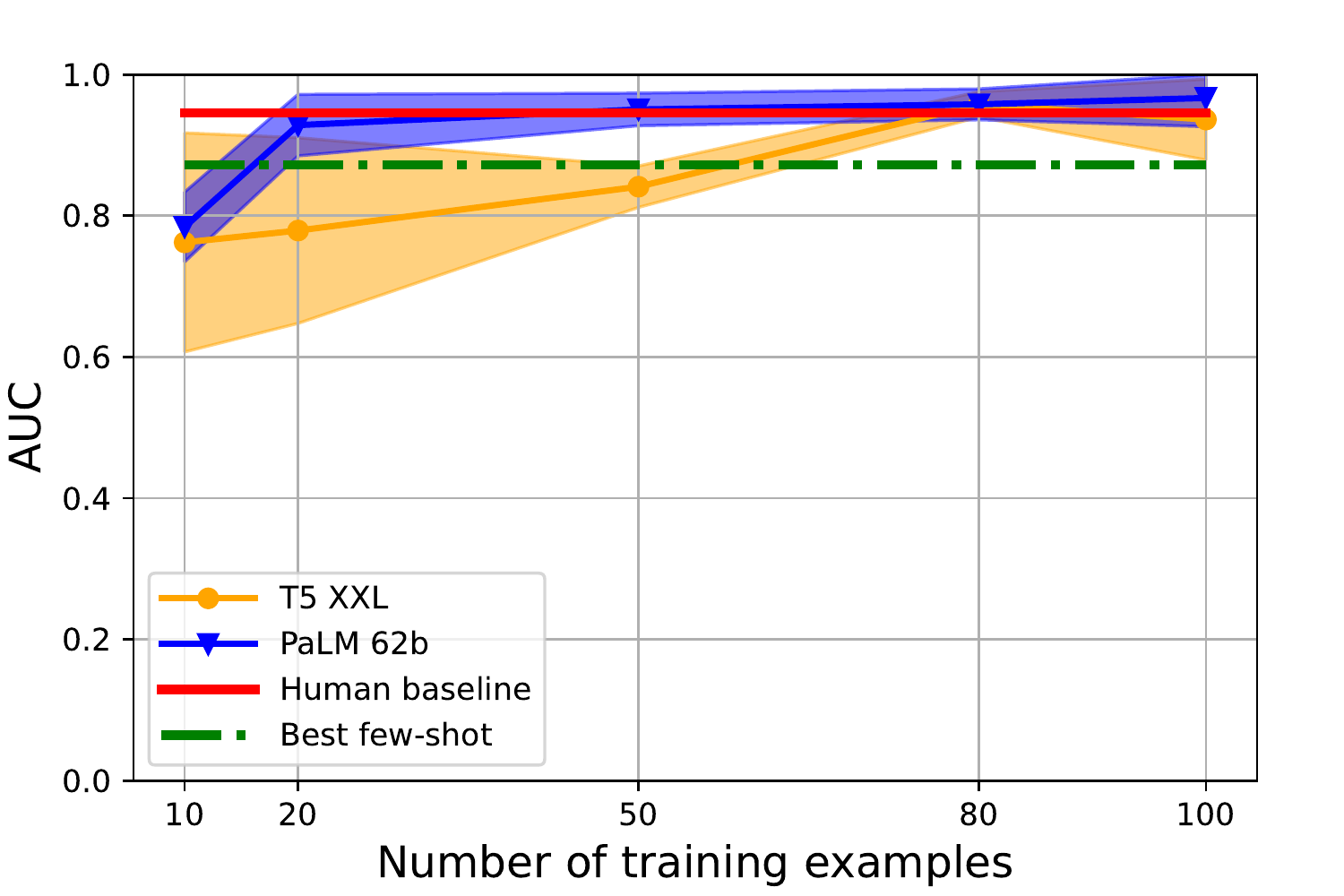}
         \caption{Neutral}
     \end{subfigure}
     \hfill
    \caption{Prompt-tuning results (ROC-AUC) on the two remaining attributes of the Neutral Responses dataset, using \tfivexxl{} and \palm{}. For few-shot, we show \palm{}'s best performing score across 0, 6, and 12 shots.}
    \label{fig:linguistic_neutrality}
\end{figure*}

The prompt-tuning results on the attributes \textit{multiple perspectives} and \textit{neutral} can be found in Figure~\ref{fig:linguistic_neutrality}. As can be seen, for \palm{}, 20 training examples are enough to outperform the Human baseline for the \textit{multiple perspectives} attribute, whereas for the \textit{neutral} one around 80 examples are needed. Furthermore, prompt-tuning \palm{} performs better than the few-shot baselines for the majority of comparisons.

Looking at \tfivexxl{}, we observe that the model does on average perform worse than \palm{} and needs 80 training examples to perform on par with the Human baseline. Nevertheless, we can see that prompt-tuning \tfivexxl{} on 20 examples suffices to outperform the \palm{} few-shot baseline for \textit{multiple perspectives} (80 examples are needed for \textit{neutral}).

Overall, these results underline the effectiveness of prompt-tuning on small datasets.\\\\

\section{Computational budget}
We estimate that it took around 1075200 GPU hours in total to create this research paper. The cost of reproducing our final results would be around 76800 GPU hours. Our model training times were in the range of 1 to 4 hours.

\end{document}